\documentclass[]{fairmeta}
\usepackage{enumitem}

\title{Recurrent Harness for Agentic Continual Learning}

\author[1,*]{Yuxin Xiao}
\author[2]{Zhiyun Lu}
\author[2]{Jianyu Wang}
\author[3]{Yangtian Zhang}
\author[2]{Yang Song}
\author[2]{Qifan Wang}
\author[2]{Bailin Wang}

\affiliation[1]{Massachusetts Institute of Technology}
\affiliation[2]{Meta MRS Research}
\affiliation[3]{Yale University}

\contribution[*]{Work done during an internship at Meta MRS Research.}

\abstract{Language model agents increasingly operate over long interaction streams, requiring online adaptation to distribution shifts through continual learning.
However, existing agent harnesses typically either evolve their states only through forward information flow or rely on additional replay or validation data for harness optimization.
To address this, we propose ReHarness, a recurrent harness framework for agentic continual learning that complements forward read and write with backward revision under a strictly online streaming protocol.
We ask what information could have been captured differently in the harness state to better inform the current instance in the stream.
Specifically, after each instance, we leverage the resulting trajectory as feedback to revise the reflections retrieved for that instance, thereby propagating feedback backward through the textual harness state.
Across three continual learning benchmarks and three language model backbones, ReHarness consistently outperforms strong raw-context baselines, memory- and reflection-based methods, and prompt optimization approaches.
Further analysis shows that ReHarness produces source-grounded and meaningful revisions, corrects overgeneralized or outdated experience under distribution shifts, and improves both stability and plasticity during continual learning.}

\date{\today}
\correspondence{Yuxin Xiao at \email{yuxin102@mit.edu} and Bailin Wang at \email{bailin@meta.com}}

\begin{document}

\maketitle

\section{Introduction} \label{sec:1}

Language model agents~\citep{guo2024large, wang2024survey} increasingly move beyond isolated tasks to operate over long interaction streams~\citep{ai2026memorybench, asawa2026continual, he2026benchmarking, zhu2026edgebench}.
In these settings, agents must adapt to distribution shifts while retaining useful experience from earlier instances in those streams.
Recent agent harnesses~\citep{guo2026question, ning2026code} enable such continual learning~\citep{wang2024comprehensive, shi2025continual, zheng2025towards} by maintaining persistent runtime scaffolding around a fixed large language model (LLM), often through memory~\citep{chhikara2025mem0, zhang2025a, dorovatas2026position}, reflection~\citep{lingam2025enhancing, yao2026parammem, zhang2026agentic}, and prompts~\citep{agrawal2026gepa}.
This allows agents to improve online without updating model parameters.

However, existing work typically evolves the harness state only through forward information flow.
After each instance, these systems extract useful information~\citep{chhikara2025mem0, yang2026plugmem, zhang2026memevolve} or reflect on the resulting trajectory~\citep{lingam2025enhancing, yao2026parammem, zhang2026agentic}, integrate the experience into a stateful harness, and later retrieve from it~\citep{wu2026remember, liu2026simplemem, zhang2026adamem} to guide future decisions.
Nevertheless, once written, memories and reflections are rarely updated based on their effectiveness on later instances.
While some methods~\citep{chhikara2025mem0, suzgun2026dynamic, zhang2026agentic} curate the harness state through operations such as add and delete, these operations primarily maintain a coherent memory or reflection bank rather than incorporate feedback from downstream use.
Other methods~\citep{cai2026from, kang2026harness, lee2026metaharness, zhang2026memevolve} instead optimize the harness state using additional training or validation sets, which are unavailable under a strictly online streaming protocol.
Under continual distribution shifts, outdated memories or overconfident reflections can therefore persist in the harness state and repeatedly misguide future decisions.

To address this gap, we propose ReHarness to close the recurrent loop for agentic continual learning.
ReHarness complements the forward information flow in existing work with a backward revision mechanism.
The revision asks what information from past experience could have been captured differently to better inform the current instance.
Specifically, after each instance, we use a revision agent to revisit the reflections retrieved for that instance.
We use the resulting trajectory as feedback for revision, while requiring any new content to remain grounded in the reflections' source trajectories.
In this way, ReHarness propagates feedback backward through the textual harness state and carries the revised state forward to future instances.
This enables online harness adaptation under a strictly streaming protocol without instance replay or gradient-based parameter updates.
Together, forward read and write, coupled with backward revision, form a recurrent harness that can promptly correct outdated memories or overconfident reflections under distribution shifts.

We demonstrate that ReHarness consistently improves continual learning across three benchmarks~\citep{ai2026memorybench, asawa2026continual, he2026benchmarking} and three backbone LLMs~\citep{anthropic2026claude48opus, google2026gemini37flash, openai2026gpt55}.
In particular, ReHarness outperforms strong raw-context baselines, structured memory and reflection methods~\citep{chhikara2025mem0, lingam2025enhancing, ouyang2026reasoningbank, zhang2026agentic}, and prompt optimization approaches~\citep{agrawal2026gepa}.
These improvements are statistically significant against all baselines, and the gains remain broadly consistent across the 16 evaluated tasks, including challenging settings where most baselines degrade relative to their stateless backbones.
Further analysis confirms that ReHarness produces source-grounded, meaningful revisions and improves both stability and plasticity during continual learning.
Qualitative examples and harness state dynamics illustrate how ReHarness corrects early overgeneralizations and revises outdated experience after distribution shifts.
Finally, ablation studies show that ReHarness does not require a large revision set to outperform the strongest full-history baseline, while using roughly one-third as many tokens.

We summarize our contributions as follows:
\begin{itemize}[noitemsep,topsep=0pt,leftmargin=*]
    \item We propose ReHarness, which combines forward information flow with backward revision to form a recurrent harness for agentic continual learning in an online streaming setting.
    \item We show that ReHarness outperforms strong baselines across three continual learning benchmarks and three LLM backbones, with statistically significant gains consistent across challenging tasks.
    \item We demonstrate that ReHarness produces source-grounded and meaningful revisions, adapts overgeneralized or outdated reflections under distribution shifts, and improves stability and plasticity.
\end{itemize}
\section{Related Work} \label{sec:2}

\paragraph{Continual Learning.}
Continual learning~\citep{wang2024comprehensive, shi2025continual, zheng2025towards} aims to adapt models to evolving data streams under distribution shifts while preventing catastrophic forgetting.
For LLMs, recent work has extended this setting to test-time adaptation~\citep{akyurek2025the, hu2025testtime, zuo2025ttrl, feng2026inplace} and self-evolving systems~\citep{tao2024survey, gao2026a} that improve from interaction feedback.
In agentic settings, learning signals are often accumulated through memory~\citep{zhang2025a, dorovatas2026position, luo2026storage}, prompts~\citep{agrawal2026gepa}, or persistent runtime states~\citep{park2023generative, karten2026continual}, enabling online adaptation while keeping the underlying LLM parameters fixed.
Recent benchmarks~\citep{shridhar2021alfworld, zhou2024webarena, ai2026memorybench, asawa2026continual, harrington2026does, he2026benchmarking} evaluate such systems over long interaction streams, measuring both plasticity to distribution shifts and stability of acquired capabilities.
In this work, we study agentic continual learning under a strictly streaming protocol and improve online adaptation in memory space by revising accumulated experience based on its downstream effectiveness.

\paragraph{Agent Harness.}
Agent harnesses~\citep{guo2026question, ning2026code} provide runtime scaffolding around a fixed LLM, turning text generation into goal-directed, multi-step behavior.
A major line of work on agent harnesses for continual learning focuses on memory, including how raw experiences are stored~\citep{packer2023memgpt, chhikara2025mem0, wang2026lightwm, rajesh2026panini}, retrieved~\citep{liu2026evolvemem, guo2026cascade, wu2026remember, liu2026simplemem, zhang2026adamem}, and optimized~\citep{kang2026harness, karten2026continual, lee2026metaharness, li2026justintime, liu2026adaptive, ma2026memchain, pan2026retrospective, xu2026test, zhang2026memevolve, zhang2026self} to support future instances.
Some methods optimize memory offline~\citep{cai2026from, ye2026umem}, while others perform online in-context learning~\citep{zheng2024synapse, sarukkai2025selfgenerated} by accumulating past demonstrations.
More recent approaches instead abstract experience into high-level reflections~\citep{lingam2025enhancing, suzgun2026dynamic, ouyang2026reasoningbank, yao2026parammem} and manage them through curation~\citep{li2026textual, wu2026forget, yan2026memory, yang2026plugmem, zhang2026agentic}.
However, existing approaches typically either rely on additional validation sets or trajectory replay for harness evolution, or support only forward information flow without adapting previously learned experience based on its downstream effectiveness.
Our work closes this recurrent loop in an online setting by complementing forward accumulation with backward revision.
\begin{figure}[t!]
    \centering
    \includegraphics[width=\textwidth]{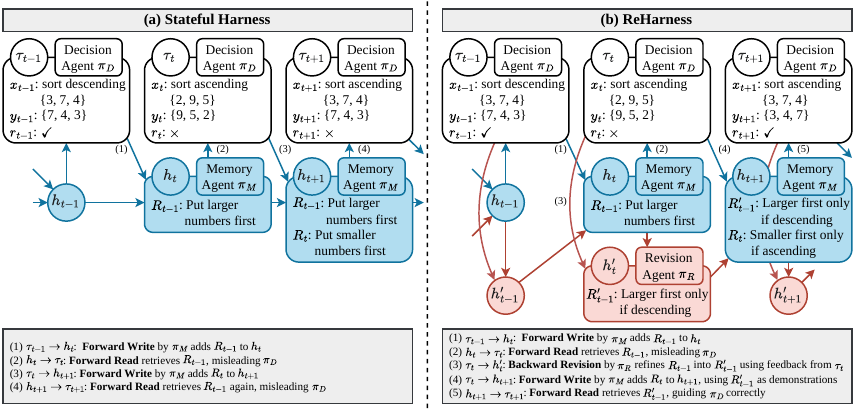}
    \caption{From stateful agent harnesses to ReHarness.
    (a) Traditional stateful harness propagates information through forward write and read, allowing an overgeneralized reflection to persist and mislead future instances.
    (b) ReHarness adds backward revision to refine retrieved reflections using downstream feedback, preventing the same error from propagating to future instances.}
    \label{fig:pipeline}
\end{figure}

\section{Framework} \label{sec:3}

We first motivate recurrent agent harnesses by identifying limitations of forward-only harness evolution in \S\ref{sec:3.1}, then introduce textual backward feedback through revision to address the challenges of agentic continual learning in \S\ref{sec:3.2}, and finally propose the ReHarness framework in \S\ref{sec:3.3}.

\subsection{From Stateful to Recurrent Agent Harnesses} \label{sec:3.1}

An agent harness for continual learning~\citep{ai2026memorybench, asawa2026continual, dorovatas2026position, he2026benchmarking} provides stateful runtime scaffolding around a decision agent $\pi_D$ over a stream of instances.
After observing the previous instance trajectory $\tau_{t-1}$, a memory agent $\pi_M$ produces a reflection $R_{t-1}$ and writes it into the current harness state $h_t$.
When the next instance arrives, the harness retrieves relevant reflections from $h_t$ to assist $\pi_D$, which responds to query $x_t$ with $y_t$ and receives reward $r_t$ from the environment.
We denote the resulting trajectory of the current instance as $\tau_t=(x_t,y_t,r_t)$.
This process repeats throughout the stream $\{\tau_t\}_{t=1}^T$.
Existing work~\citep{zhao2024expel, lingam2025enhancing, ouyang2026reasoningbank, zhang2026agentic} thus evolves the harness state through alternating forward write ($\tau_{t-1} \rightarrow R_{t-1} \rightarrow h_t$) and forward read ($h_t \rightarrow R_{t-1} \rightarrow \tau_t$).

Figure~\ref{fig:pipeline} (a) illustrates this forward information flow in stateful harnesses with a simple sorting example.
After correctly sorting $\{3,7,4\}$ in descending order at $\tau_{t-1}$, $\pi_M$ writes the reflection ``put larger numbers first'' into $h_t$.
At $\tau_t$, however, the query asks for ascending order.
Retrieving $R_{t-1}$ therefore misleads $\pi_D$, which produces an incorrect answer.
Although $\pi_M$ then writes a new reflection $R_t$ recommending ``put smaller numbers first,'' the misleading $R_{t-1}$ remains in $h_{t+1}$ and is retrieved again for $\tau_{t+1}$ because the queries contain overlapping numbers.
Consequently, the same overgeneralized reflection misguides $\pi_D$ again.

This example shows that a stateful harness with only the forward path is incomplete for continual learning.
This forward-only reflection process does not account for the effectiveness of reflections when reused in future instances under distribution shifts.
Misleading reflections can persist in the harness state and propagate errors through later retrievals.
We instead propose a \emph{recurrent} agent harness that combines forward information flow with backward feedback from future instances.

\subsection{Textual Backward Feedback through Revision} \label{sec:3.2}

Agentic continual learning under a strictly streaming setting poses unique challenges for backward feedback.
First, each instance is observed only once, so we cannot return to past instances to collect additional counterfactual outcomes for offline optimization~\citep{cai2026from, li2026justintime, pan2026retrospective}.
Second, we do not have additional validation sets to select among candidate harness updates~\citep{kang2026harness, lee2026metaharness, zhang2026self}.
Third, frontier API-based agents~\citep{anthropic2026claude48opus, google2026gemini37flash, openai2026gpt55} do not expose model parameters, precluding gradient-based updates.
On the other hand, prior work~\citep{pryzant2023automatic, shinn2023reflexion, yuksekgonul2024textgrad} shows that natural language feedback can provide an effective learning signal without parameter updates.
We therefore realize backward feedback in the textual harness state, allowing the harness to adapt online as the interaction stream evolves.

Specifically, we introduce a revision agent $\pi_R$ that revisits reflections used for the current instance based on their observed effectiveness.
Using the same sorting example in Figure~\ref{fig:pipeline} (b), after observing the current instance trajectory $\tau_t$, $\pi_R$ revises $R_{t-1}$ in to $R'_{t-1}$ using its source trajectory $\tau_{t-1}$ and the downstream evidence from $\tau_t$.
The revision aims to produce a reflection that, if used in place of $R_{t-1}$, could have prevented mistakes in $\tau_t$.
This typically involves incorporating previously missed evidence from $\tau_{t-1}$ or correcting an exaggerated hypothesis in $R_{t-1}$.
Here, $\pi_R$ adds a limiting condition to $R'_{t-1}$ by specifying ``larger first only if descending.''
Replacing $R_{t-1}$ with $R'_{t-1}$ yields a revised harness state $h'_t$, establishing the backward feedback path $\tau_t \rightarrow R'_{t-1} \rightarrow h'_t$.
Meanwhile, we use $R'_{t-1}$ as a demonstration for the forward write, which generates a new reflection $R_t$ from $\tau_t$ with a similar limiting condition: ``smaller first only if ascending.''
Combining the backward revision with this forward reflection yields the next harness state $h_{t+1} = h'_t \oplus R_t$, completing the recurrent update.
Consequently, retrieving $R'_{t-1}$ from $h_{t+1}$ helps $\pi_D$ answer $\tau_{t+1}$ correctly.

\subsection{ReHarness: Closing the Recurrent Loop} \label{sec:3.3}

We now instantiate this recurrent harness framework as ReHarness in Figure~\ref{fig:pipeline} (b).

\paragraph{Step 1: Forward Read.}
Let $h_t = \{R_i\}_{i<t}$ denote the current harness state.
Given query $x_t$, we provide the decision agent $\pi_D$ with the top-$k$ relevant reflections retrieved from $h_t$:
\begin{equation} \label{eq:read}
    \mathcal{I}_t = \{i_1,\dots,i_k\} = \operatorname{Retrieval}(x_t, h_t, k),
    \qquad
    y_t = \pi_D \! \left( x_t, \{R_i\}_{i\in\mathcal{I}_t} \right).
\end{equation}
The environment then provides reward $r_t$, forming the instance trajectory $\tau_t=(x_t,y_t,r_t)$.

\paragraph{Step 2: Backward Revision.}
After observing $\tau_t$, the revision agent $\pi_R$ independently revisits each retrieved reflection using its source trajectory $\tau_i$ and its effectiveness on the current instance $\tau_t$:
\begin{equation} \label{eq:revision}
    R'_i = \pi_R \! \left( R_i, \tau_i; \tau_t \right),
    \qquad
    \forall i \in \mathcal{I}_t.
\end{equation}
Importantly, $\tau_t$ serves only as feedback for assessing and revising $R_i$.
Any new content in $R'_i$ must remain grounded in its source trajectory $\tau_i$ to prevent information leakage.
The revision may therefore incorporate evidence missed in $\tau_i$, correct unsupported hypotheses in $R_i$, or refine the conditions under which the reflection should apply.
Replacing the retrieved reflections with their revised versions while retaining all unretrieved reflections yields the intermediate state $h_t'$.

\paragraph{Step 3: Forward Write.}
Finally, the memory agent $\pi_M$ reflects on $\tau_t$ to produce a new reflection:
\begin{equation} \label{eq:write}
    R_t = \pi_M \! \left( \tau_t; \{R'_i\}_{i\in\mathcal{I}_t} \right),
    \qquad
    h_{t+1} = h_t' \oplus R_t.
\end{equation}
We prompt $\pi_M$ to explore operational evidence and retain only well-supported interpretations or strategies.
The revised reflections $\{R'_i\}_{i\in\mathcal{I}_t}$ serve as demonstrations for $\pi_M$, while the prompt encourages $R_t$ to capture complementary perspectives not covered by those demonstrations.
\begin{table}[t!]
    \caption{Mean normalized gain across task streams for each benchmark--backbone combination. 
    Highest gains are in bold.
    ReHarness outperforms all baselines, suggesting that backward revision of accumulated experience effectively adapts agent harnesses to continual distribution shifts.}
    \centering
\resizebox{\textwidth}{!}{
    \begin{tabular}{c|ccc|ccc|ccc|c}
        \toprule
            \multirow{2}{*}{\textbf{Method}} & \multicolumn{3}{c|}{\textbf{CL-Bench}} & \multicolumn{3}{c|}{\textbf{MemoryBench}} & \multicolumn{3}{c|}{\textbf{MemoryArena}} & \multirow{2}{*}{\textbf{Average}} \\
            & \textbf{Claude} & \textbf{GPT} & \textbf{Gemini} & \textbf{Claude} & \textbf{GPT} & \textbf{Gemini} & \textbf{Claude} & \textbf{GPT} & \textbf{Gemini} \\
        \midrule
        \midrule
             \textbf{Mem0} & $0.103$ & $0.220$ & $0.092$ & $0.049$ & $0.016$ & $0.053$ & $0.093$ & $0.322$ & $0.235$ & $0.131$ \\
             \textbf{RBank} & $0.073$ & $0.065$ & $0.040$ & $0.023$ & $0.004$ & $0.045$ & $0.014$ & $0.017$ & $0.023$ & $0.034$ \\
             \textbf{DoT} & $0.125$ & $0.081$ & $-0.008$ & $0.053$ & $0.027$ & $0.010$ & $0.129$ & $0.179$ & $0.042$ & $0.071$ \\
             \textbf{ACE} & $0.042$ & $0.088$ & $0.036$ & $0.085$ & $-0.018$ & $0.028$ & $0.045$ & $0.004$ & $0.073$ & $0.043$ \\
             \textbf{GEPA-Online} & $0.076$ & $0.239$ & $0.132$ & $0.138$ & $0.064$ & $-0.134$ & $0.126$ & $0.129$ & $0.058$ & $0.092$ \\
             \textbf{Top-K Raw} & $0.165$ & $0.268$ & $0.190$ & $0.035$ & $0.041$ & $0.037$ & $0.145$ & $0.340$ & $0.291$ & $0.168$ \\
             \textbf{Full Raw} & $0.191$ & $0.299$ & $\mathbf{0.234}$ & $0.063$ & $0.024$ & $-0.013$ & $0.082$ & $\mathbf{0.358}$ & $0.237$ & $0.164$ \\
             \textbf{ReHarness} & $\mathbf{0.217}$ & $\mathbf{0.317}$ & $0.233$ & $\mathbf{0.160}$ & $\mathbf{0.096}$ & $\mathbf{0.064}$ & $\mathbf{0.305}$ & $\mathbf{0.358}$ & $\mathbf{0.408}$ & $\mathbf{0.240}$ \\
        \bottomrule
    \end{tabular}
}
    \label{tab:main}
\end{table}

\section{Experiment} \label{sec:4}

\subsection{Experiment Setup} \label{sec:4.1}

\paragraph{Baselines.}
We compare ReHarness against the following strong baselines.
\begin{itemize}[noitemsep,topsep=0pt,leftmargin=*]
    \item \textbf{Mem0}~\citep{chhikara2025mem0} maintains atomic memories as harness state, retrieving top-$k$ entries for the decision agent and incrementally extracting memories after each instance.
    \item \textbf{RBank}~\citep{ouyang2026reasoningbank} maintains an append-only bank of reusable reflections as harness state and retrieves top-$k$ reflections for each instance.
    \item \textbf{DoT}~\citep{lingam2025enhancing} generates diverse reflections from the top-$k$ retrieved prior experiences and uses them as the runtime harness for the current instance.
    \item \textbf{ACE}~\citep{zhang2026agentic} maintains a global reflection playbook as its harness state and curates it through add, update, delete, and merge operations for coherence.
    \item \textbf{GEPA-Online} adapts GEPA~\citep{agrawal2026gepa} to our strictly online streaming setting, preserving reflective optimization of the decision prompt as persistent harness state, but replaces the original validation-based population search with sequential updates from each observed trajectory.
    \item \textbf{Full Raw} uses all prior instance trajectories in the stream as harness context, while \textbf{Top-K Raw} retrieves only the $k$ most relevant trajectories for in-context learning.   
\end{itemize}
We evaluate all methods with three frontier LLM backbones: Claude Opus 4.8~\citep{anthropic2026claude48opus}, GPT-5.5~\citep{openai2026gpt55}, and Gemini 3.7 Flash~\citep{google2026gemini37flash}, using the same backbone for all agents within each method.
We use each model's default generation configuration with high thinking effort and exclude hidden reasoning traces from all harness operations.
Retrieval uses the top five entries ranked by Qwen3-Embedding-8B~\citep{zhang2025qwen3} and applies the same query-to-query comparison across all methods.
We provide the prompts used by ReHarness in \S\ref{app:a}.

\paragraph{Datasets.}
We evaluate ReHarness against baselines on three continual learning benchmarks under a strictly online streaming protocol, with no training or validation set and no repeated trials.
\begin{itemize}[noitemsep,topsep=0pt,leftmargin=*]
    \item \textbf{CL-Bench}~\citep{asawa2026continual} contains six tasks spanning cohort studies, codebase adaptation~\citep{jimenez2024swebench}, database exploration~\citep{hou2026bridging}, sales prediction, blind spectrum monitoring, and exploitable poker.
    Each task contains 12 to 120 multi-round instances with explicit distribution shifts and task-specific evaluation metrics.
    \item \textbf{MemoryBench}~\citep{ai2026memorybench} contains long-input, long-output generation tasks across multiple domains.
    We use its five English tasks from HelloBench~\citep{que2024hellobench} and WritingBench~\citep{wu2026writingbench}, covering academic, creative, and political writing, with 17 to 47 single-round instances per task, evaluated by GPT-5.6-Sol~\citep{openai2026gpt56sol}.
    \item \textbf{MemoryArena}~\citep{he2026benchmarking} contains five domains spanning travel planning~\citep{xie2024travelplanner}, web shopping~\citep{yao2022webshop}, progressive web search~\citep{chen2026browsecomp}, and formal reasoning in mathematics and physics, with 20 to 270 tasks per domain.
    Each task comprises 2 to 16 interdependent instances, each involving multiple interaction rounds.
\end{itemize}
Across these benchmarks, we standardize each task as a continual learning stream and each constituent interaction unit as an instance (i.e., $\tau_t$).
ReHarness and all baselines maintain harness state across instances within a stream and reset it between streams.
To compare performance across tasks with different native metrics, we follow~\citet{asawa2026continual} and report normalized gain,
$(r_\text{method} - r_\text{stateless}) / (r_{\max} - r_\text{stateless})$,
averaged over each stream.
Here, $r_{\max}$ denotes the maximum achievable reward under each task's native metric.
This metric quantifies the improvement over the stateless backbone agent from learning prior experience within the stream.
We repeat each experiment over three independently sampled stream rollouts.
Within each rollout, all methods share the same benchmark initialization and random seed, then evolve independently.

\begin{figure}[t!]
    \centering
    \includegraphics[width=\textwidth]{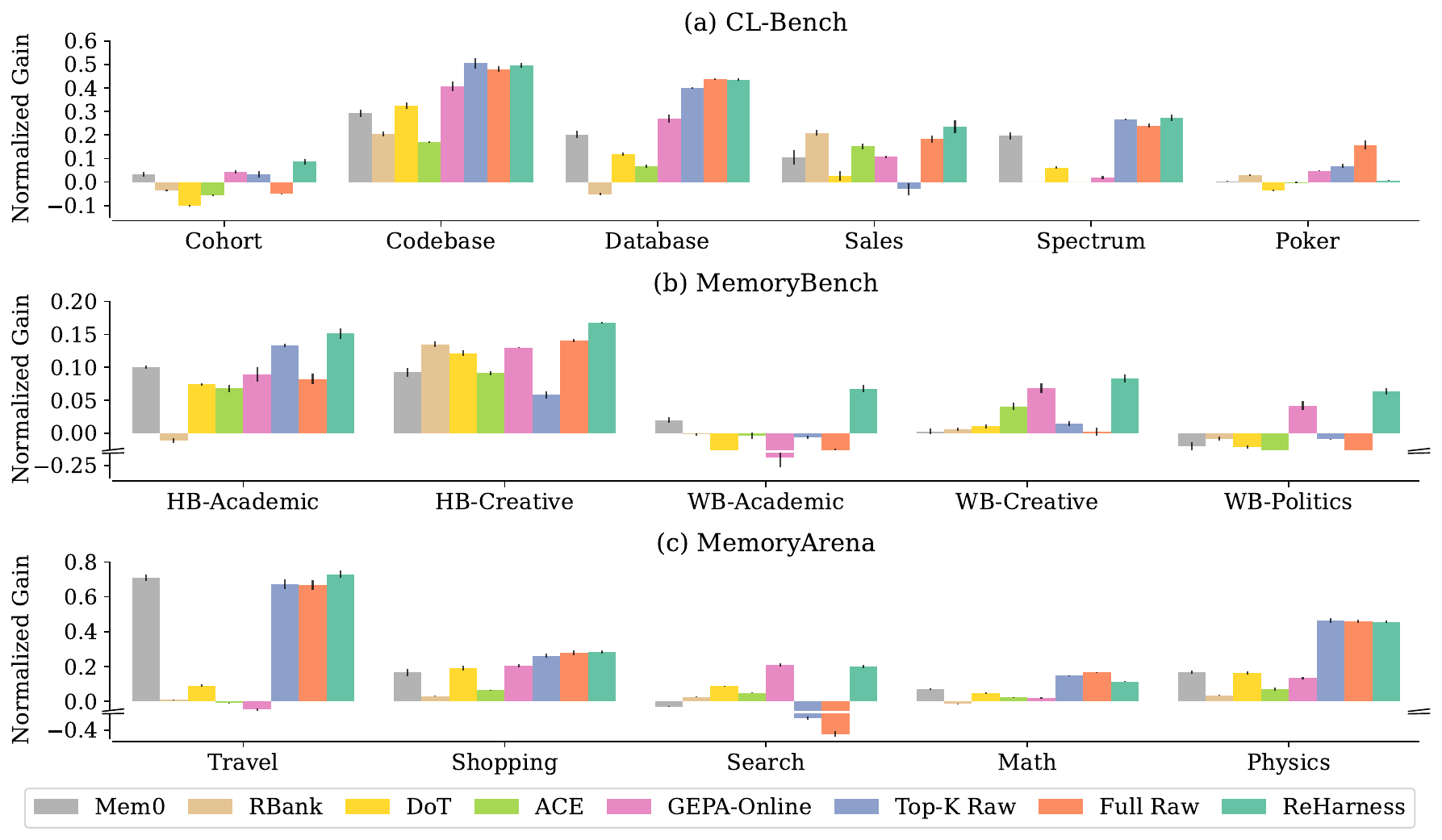}
    \caption{Task-level normalized gain for each method, averaged across the three backbone LLMs.
    Performance varies across tasks, while ReHarness remains effective on nearly all of them.}
    \label{fig:task}
\end{figure}

\subsection{ReHarness Consistently Improves Agentic Continual Learning} \label{sec:4.2}

\paragraph{Overall Performance.}
Table~\ref{tab:main} reports the mean normalized gain across task streams for each benchmark--backbone combination.
ReHarness achieves the highest overall normalized gain of $0.240$ and ranks first or tied in eight of the nine combinations.
Consistent with~\citet{asawa2026continual}, raw-context methods are strong baselines: Top-K Raw and Full Raw outperform most structured memory and reflection methods on average, particularly on CL-Bench and MemoryArena.
Reflection-based methods, such as RBank, DoT, and ACE, are less consistent, with some settings yielding negative gains.
Similarly, GEPA-Online degrades substantially on MemoryBench with Gemini, suggesting that prompt optimization alone is less robust for continual learning.
To assess statistical significance, we compare ReHarness against each baseline using two-sided paired Wilcoxon signed-rank tests with Bonferroni correction, treating each seeded task stream as one observation.
ReHarness significantly outperforms all baselines ($p_{\mathrm{adj}}<0.05$), including the strong raw-context baselines Top-K Raw ($p_{\mathrm{adj}}=0.039$) and Full Raw ($p_{\mathrm{adj}}=0.043$).
Overall, these results suggest that revising reflections with ReHarness is more effective than retaining raw history or relying on forward-only accumulation.

\paragraph{Task-Level Results.}
We next present the task-level breakdown for each benchmark in Figure~\ref{fig:task}, averaged across the three backbone LLMs.
The gains vary substantially across tasks, with many methods showing marginal or even negative improvements over their stateless backbones.
On CL-Bench, ReHarness leads or remains close to the best on most tasks, but shows only marginal gains on exploitable poker.
On MemoryBench, while most methods perform well on HelloBench (HB-Academic and HB-Creative), only ReHarness achieves consistent gains across WritingBench (WB-Academic, WB-Creative, and WB-Politics).
On MemoryArena, Mem0 is notably strong on travel planning, and GEPA-Online performs well on progressive web search, while ReHarness remains consistently competitive across all five domains.
A closer examination suggests that, unlike the more deterministic codebase and database settings, exploitable poker is more stochastic: hand-level rewards can be weakly coupled with strategy quality due to hidden cards and chance.
Raw-context baselines are less exposed to this credit-assignment issue because they preserve the original hand trajectories and defer their interpretation to the decision agent at each new hand.
By contrast, ReHarness is less effective when downstream rewards provide a noisy signal for backward revision, and benefits most when feedback is diagnostic for revising reusable reflections.
The full raw results are in \S\ref{app:a}.

\paragraph{Token Efficiency.}
We further examine the token consumption and API cost of ReHarness.
Despite requiring one reflection call for forward write and $k$ revision calls for backward revision, in addition to the decision calls shared by all methods, ReHarness consumes fewer total tokens than the two strong raw-context baselines.
This is because compact reflections substantially reduce the repeated decision-time context cost of raw trajectories, especially for long, multi-round instances.
Across one rollout over all CL-Bench tasks with Claude, ReHarness consumes ${\sim}37.19$M input and ${\sim}4.38$M output tokens, totaling ${\sim}41.57$M.
By comparison, Top-K Raw consumes ${\sim}52.20$M input and ${\sim}0.46$M output tokens (${\sim}52.66$M total), while Full Raw consumes ${\sim}141.68$M input and ${\sim}0.44$M output tokens (${\sim}142.12$M total), or $1.3\times$ and $3.4\times$ as many total tokens as ReHarness, respectively.
Thus, although ReHarness generates more output tokens, its input-token savings more than offset this overhead in total consumption.
With prompt caching and Claude API pricing, the corresponding costs are approximately $\$85$ for Top-K Raw, $\$254$ for ReHarness, and $\$314$ for Full Raw.
ReHarness therefore remains substantially cheaper than Full Raw while achieving a higher normalized gain ($0.217$ vs.\ $0.191$), whereas Top-K Raw is cheaper but achieves a lower gain ($0.165$).
\section{Analysis} \label{sec:5}

\subsection{Is ReHarness Revision Source-Grounded?} \label{sec:5.1}

\begin{table}[t!]
    \caption{Average grounded, leakage, and unsupported claim rates of revised reflections on CL-Bench for each backbone, together with the average change per revision.
    The grounded rate remains above $95\%$ for all three LLMs and decreases only marginally after revision, suggesting that ReHarness uses the downstream trajectory primarily as feedback rather than as a source of leaked information.}
    \centering
\resizebox{0.8\textwidth}{!}{
    \begin{tabular}{c|cc|cc|cc}
        \toprule
            \multirow{2}{*}{\textbf{LLM}} & \multicolumn{2}{c|}{\textbf{Grounded Rate} $\mathbf{(\uparrow)}$} & \multicolumn{2}{c|}{\textbf{Leakage Rate} $\mathbf{(\downarrow)}$} & \multicolumn{2}{c}{\textbf{Unsupported Rate} $\mathbf{(\downarrow)}$} \\
            & \textbf{Average} & $\mathbf{\Delta}$ / \textbf{Revision} & \textbf{Average} & $\mathbf{\Delta}$ / \textbf{Revision} & \textbf{Average} & $\mathbf{\Delta}$ / \textbf{Revision} \\
        \midrule
        \midrule
            \textbf{Claude} & $0.953$ & $-0.003$ & $0.021$ & $0.006$ & $0.026$ & $0.002$ \\
            \textbf{GPT} & $0.987$ & $-0.002$ & $0.008$ & $0.003$ & $0.005$ & $0.001$ \\
            \textbf{Gemini} & $0.963$ & $0.007$ & $0.013$ & $0.004$ & $0.024$ & $-0.002$ \\
        \bottomrule
    \end{tabular}
}
    \label{tab:grounding}
\end{table}

We first examine whether backward revision in Equation~\ref{eq:revision} remains grounded in its source trajectory $\tau_i$ or imports information from $\tau_t$.
Following~\citet{min2023factscore}, we decompose each revised reflection $R'_i$ into factual claims using GPT-5.6-Sol~\citep{openai2026gpt56sol}.
We classify each claim as: (1) grounded, if entailed by $\tau_i$; (2) leaked, if not entailed by $\tau_i$ but entailed by $\tau_t$; or (3) unsupported, if entailed by neither.
Table~\ref{tab:grounding} reports the average rates across CL-Bench tasks for each backbone and their average change per revision.
Across all backbones, over $95\%$ of claims remain source-grounded, with leakage at $0.8$--$2.1\%$ and unsupported claims at $0.5$--$2.6\%$.
Moreover, revision changes these rates only marginally, with grounding decreasing by at most $0.3\%$ per revision.
These results indicate that the backbone LLMs largely follow the grounding constraint as intended, and ReHarness uses $\tau_t$ primarily as feedback rather than as a source of leaked information.

\subsection{Is ReHarness Capable of Revision?} \label{sec:5.2}

We next validate whether ReHarness revises the harness state as intended.
The goal of revision is to use downstream feedback to improve previously learned experience for future instances.
We therefore consider a counterfactual replay that modifies the online streaming protocol.
After revising the harness state into $h'_t$ based on the completed trajectory $\tau_t$, we reuse $h'_t$ to assist the decision agent on the same instance.
This differs from ReHarness, which applies $h'_t$ only to the next streamed instance $\tau_{t+1}$.
Importantly, $h'_t$ remains grounded in past trajectories and uses $\tau_t$ only as feedback on their effectiveness.
The replay thus asks: given the current outcome, what information already available in past trajectories should have been captured differently to better inform that instance?

Table~\ref{tab:counterfactual} reports task-level normalized gains on CL-Bench with Claude for ReHarness, counterfactual replay, and a no-revision baseline that removes the revision agent $\pi_R$ and retains only forward information flow.
Counterfactual replay substantially improves over ReHarness across all tasks.
This suggests that revision can recover or correct information from past trajectories using current feedback, producing a more effective harness state had those revisions been available beforehand.
Hence, counterfactual replay confirms that the backward mechanism produces meaningful revisions.
However, it violates the online streaming protocol.
ReHarness instead applies the revised state $h'_t$ only to the next instance $\tau_{t+1}$, using feedback from $\tau_t$ to improve future instances.
Compared with the no-revision baseline, ReHarness achieves higher gains across all tasks, especially in database exploration, sales prediction, and blind spectrum monitoring.
This indicates that backward revision remains beneficial when applied only to subsequent instances in the stream.

\begin{table}[t!]
    \caption{Normalized gain under counterfactual revision on CL-Bench with Claude.
    Counterfactual replay reapplies the revised harness to the same instance and substantially improves over ReHarness and the no-revision baseline.
    This shows that the backward mechanism enables meaningful revision.}
    \centering
\resizebox{\textwidth}{!}{
    \begin{tabular}{c|cc|cccccc|c}
        \toprule
            \textbf{Revision} & \textbf{Revision} & \textbf{Revised Harness} & \multicolumn{7}{c}{\textbf{CL-Bench}} \\
            \textbf{Setting} & \textbf{Based On} & \textbf{Applied To} & \textbf{Cohort} & \textbf{Codebase} & \textbf{Database} & \textbf{Sales} & \textbf{Spectrum} & \textbf{Poker} & \textbf{Average} \\
        \midrule
        \midrule
            \textbf{No Revision} & -- & -- & $-0.053$ & $0.359$ & $0.279$ & $0.258$ & $0.033$ & $-0.012$ & $0.144$ \\
            \textbf{ReHarness} & $\tau_t$ & $\tau_{t+1}$ & $-0.019$ & $0.378$ & $0.361$ & $0.408$ & $0.162$ & $0.009$ & $0.217$ \\
            \textbf{Counterfactual} & $\tau_t$ & $\tau_t$ & $\mathbf{0.046}$ & $\mathbf{0.509}$ & $\mathbf{0.402}$ & $\mathbf{0.698}$ & $\mathbf{0.314}$ & $\mathbf{0.041}$ & $\mathbf{0.335}$ \\
        \bottomrule
    \end{tabular}
}
    \label{tab:counterfactual}
\end{table}

\subsection{What Does ReHarness Revise during Continual Learning?} \label{sec:5.3}

\begin{figure}
    \centering
    \includegraphics[width=\linewidth]{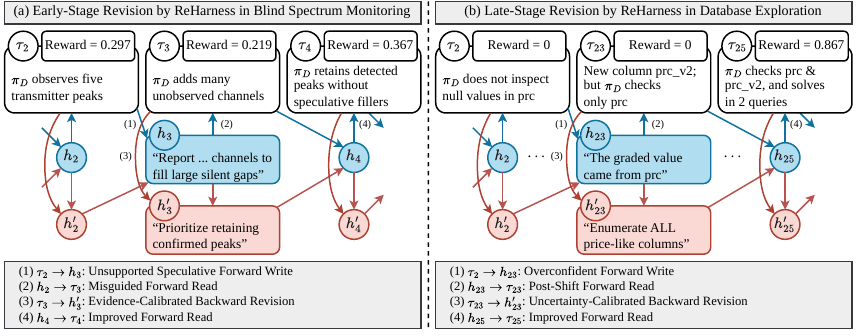}
    \caption{Qualitative examples of ReHarness revision during continual learning.
    (a) In blind spectrum monitoring, ReHarness corrects an early overgeneralized reflection with evidence-grounded guidance.
    (b) In database exploration, ReHarness adapts an outdated reflection to a schema shift.}
    \label{fig:example}
\end{figure}

Now that we have established the source grounding and capability of ReHarness revision, we use qualitative examples from CL-Bench with Claude in Figure~\ref{fig:example} to examine what it revises.
Figure~\ref{fig:example} (a) shows an early-stage example from blind spectrum monitoring, where the agent infers persistent transmitters from sequential spectrum scans.
After observing $\tau_2$, the memory agent $\pi_M$ overgeneralizes the evidence and recommends ``report[ing] plausible additional channels to fill large silent gaps.''
Because the task rewards overlap with the ground-truth occupied spectrum, both missing real transmitters and hallucinating nonexistent ones hurt performance.
When this reflection is retrieved from $h_3$ for $\tau_3$, the decision agent $\pi_D$ follows the speculative guidance, adds unobserved transmitters, and receives a lower reward.
Using this outcome as backward feedback, the revision agent $\pi_R$ marks the hypothesis as ``speculative'' and changes the guidance to ``prioritize retaining confirmed peaks over adding unconfirmed ones,'' producing $h'_3$.
When the revised reflection is retrieved from $h_4$, $\pi_D$ avoids speculative fillers and substantially improves the reward.
This example shows how ReHarness corrects an early overgeneralization before it misguides future decisions.

The second example in Figure~\ref{fig:example} (b) shows a late-stage revision from database exploration in CL-Bench, where the agent queries an unknown SQLite database.
During $\tau_2$, $\pi_D$ fails to inspect null values in a price column ``prc,'' yet $\pi_M$ eventually consolidates the reflection that ``the graded value came from prc'' after repeated use in the early stream.
After $\tau_{20}$, CL-Bench introduces a distribution shift by changing the schema and adding ``prc\_v2.''
When the old-schema reflection is retrieved from $h_{23}$ for $\tau_{23}$, $\pi_D$ checks only ``prc,'' while the correct answer lies in ``prc\_v2.''
Using this failure as backward feedback, $\pi_R$ changes the rule to ``not assume which column is graded'' and ``enumerate all price-like columns,'' producing $h'_{23}$.
When this revised reflection is retrieved for $\tau_{25}$, $\pi_D$ checks both ``prc'' and ``prc\_v2'' and obtains the correct answer in two queries, substantially improving the reward.
This example shows how ReHarness revises outdated strategies under distribution shifts through backward feedback.
\S\ref{app:a} provides the full trajectories, reflections, and revisions.

\begin{table}[t!]
    \caption{Stability--plasticity decomposition on CL-Bench.
    ReHarness achieves the highest average gains in both, showing robust transfer across distribution shifts and effective adaptation afterward.}
    \centering
\resizebox{\textwidth}{!}{
    \begin{tabular}{c|cc|cc|cc|cc}
        \toprule
            \multirow{2}{*}{\textbf{Method}} & \multicolumn{2}{c|}{\textbf{Claude}} & \multicolumn{2}{c|}{\textbf{GPT}} & \multicolumn{2}{c|}{\textbf{Gemini}} & \multicolumn{2}{c}{\textbf{Average}} \\
            & \textbf{Stability} & \textbf{Plasticity} & \textbf{Stability} & \textbf{Plasticity} & \textbf{Stability} & \textbf{Plasticity} & \textbf{Stability} & \textbf{Plasticity} \\
        \midrule
        \midrule
            \textbf{Mem0} & $0.002$ & $0.101$ & $0.014$ & $0.206$ & $-0.026$ & $0.118$ & $-0.003$ & $0.142$ \\
            \textbf{RBank} & $\mathbf{0.004}$ & $0.069$ & $0.001$ & $0.065$ & $-0.022$ & $0.062$ & $-0.006$ & $0.065$ \\
            \textbf{DoT} & $-0.003$ & $0.127$ & $-0.002$ & $0.083$ & $0.000$ & $-0.007$ & $-0.002$ & $0.068$ \\
            \textbf{ACE} & $0.000$ & $0.042$ & $0.011$ & $0.077$ & $-0.028$ & $0.065$ & $-0.006$ & $0.061$ \\
            \textbf{GEPA-Online} & $0.000$ & $0.076$ & $0.006$ & $0.223$ & $-0.017$ & $0.149$ & $-0.004$ & $0.153$ \\
            \textbf{Top-K Raw} & $-0.012$ & $0.177$ & $0.009$ & $0.260$ & $0.011$ & $0.179$ & $0.003$ & $0.205$ \\
            \textbf{Full Raw} & $-0.005$ & $0.196$ & $0.010$ & $0.290$ & $\mathbf{0.024}$ & $0.210$ & $0.010$ & $0.232$ \\
            \textbf{ReHarness} & $0.003$ & $\mathbf{0.214}$ & $\mathbf{0.025}$ & $\mathbf{0.291}$ & $0.016$ & $\mathbf{0.217}$ & $\mathbf{0.015}$ & $\mathbf{0.241}$ \\
        \bottomrule
    \end{tabular}
}
    \label{tab:decomposition}
\end{table}

\subsection{How Does ReHarness Revise during Continual Learning?} \label{sec:5.4}

We then investigate how ReHarness revises accumulated experience during continual learning.
Following~\citet{asawa2026continual}, we decompose normalized gain on CL-Bench into stability and plasticity.
Stability measures gain on the first instance after a distribution shift, before post-shift feedback is available.
Plasticity measures gain on subsequent instances before the next shift, where methods can adapt to the current distribution.
Because the two categories contain different numbers of instances, their absolute gains are not directly comparable.
As shown in Table~\ref{tab:decomposition}, Mem0, RBank, DoT, ACE, and GEPA-Online exhibit negative stability gains on average, indicating that their forward-only harness states can become detrimental under distribution shifts.
In contrast, ReHarness achieves the highest stability gain on average and the highest plasticity gain for every backbone LLM.
These results suggest that backward revision enables more robust transfer across shifts while supporting effective adaptation once post-shift feedback becomes available, as illustrated in Figure~\ref{fig:example}.

\begin{figure}[t!]
    \centering
    \includegraphics[width=\textwidth]{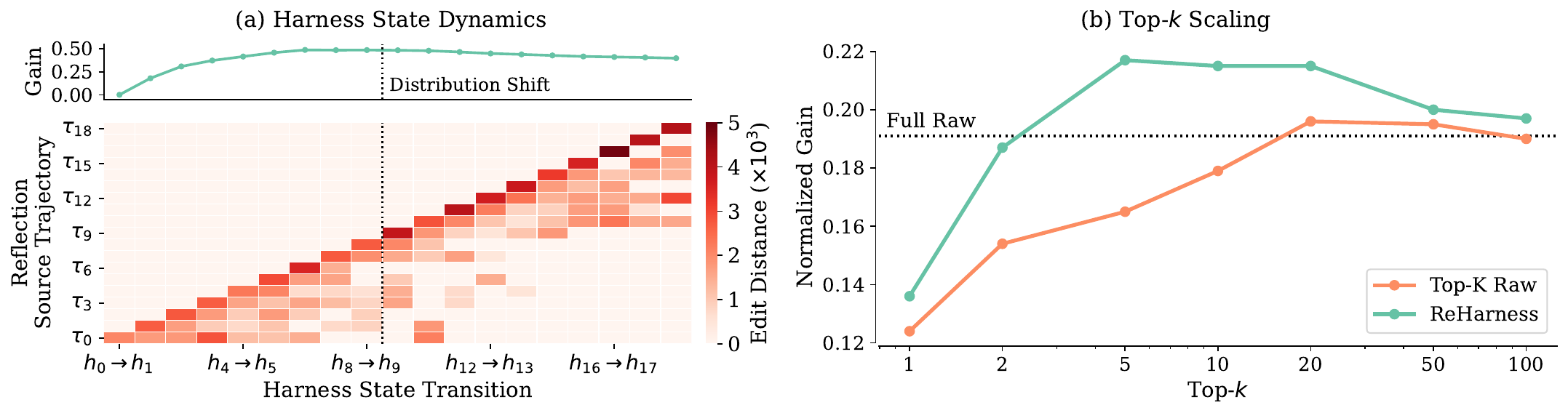}
    \caption{(a) Harness state dynamics by ReHarness on CL-Bench codebase adaptation with Claude.
    Revisions diminish as reflections stabilize before the distribution shift, then rise afterward to update outdated reflections.
    (b) Top-$k$ scaling on CL-Bench with Claude.
    ReHarness improves sharply up to $k=5$ and stabilizes for moderate $k$, while larger values yield diminishing returns and higher cost.}
    \vspace{-4pt}
    \label{fig:analysis}
\end{figure}

We further visualize harness state dynamics on CL-Bench codebase adaptation with Claude.
Each transition from $h_t$ to $h_{t+1}$ adds a new reflection $R_t$ through forward write and revises retrieved reflections $\{R'_i\}_{i\in\mathcal{I}_t}$ through backward revision.
In Figure~\ref{fig:analysis} (a), diagonal heatmap entries mark new reflections, while later non-zero entries in the same row indicate revisions.
We measure revision magnitude by character-level edit distance between the old and revised reflections from the same source trajectory.
Before the distribution shift at $\tau_9$, revisions to older reflections diminish as the harness state stabilizes and the gain rises.
After the shift, older reflections require larger revisions to adapt to the new distribution.
For example, the transition from $h_{10}$ to $h_{11}$ substantially revises reflections from $\tau_0$ and $\tau_1$.
These revisions correct outdated experience while the gain remains relatively stable after the shift.
As the stream progresses, revision increasingly concentrates on recent reflections relevant to the shifted distribution.
Together, these dynamics show how ReHarness balances stability and plasticity by preserving useful experience while revising outdated reflections.

\subsection{What Makes ReHarness Effective?} \label{sec:5.5}

Finally, we conduct ablations to identify what makes ReHarness effective.
In Figure~\ref{fig:analysis} (b), we vary the number of retrieved and subsequently revised reflections, $k$, on a log scale for ReHarness and Top-K Raw on CL-Bench with Claude.
Top-K Raw approaches Full Raw as $k$ reaches $20$, suggesting diminishing returns from additional in-context examples.
For ReHarness, normalized gain rises sharply from $k=1$ to $5$, remains relatively stable for $k\in\{5,10,20\}$, and declines for larger $k$.
This suggests that overly large $k$ can introduce unnecessary revision and potential reflection drift while increasing computational cost.
ReHarness therefore does not require a large revision set.
We fix $k=5$ throughout the main experiments and report this sweep as a post-hoc sensitivity analysis.

Furthermore, we ablate two components of ReHarness.
The effect of removing backward revision is already established in Table~\ref{tab:counterfactual}.
Removing the revised reflections $\{R'_i\}_{i\in\mathcal{I}_t}$ used as demonstrations for $\pi_M$ in Equation~\ref{eq:write} reduces the normalized gain on CL-Bench with Claude from $0.217$ to $0.191$, indicating that revised experience helps guide subsequent reflection generation.
\section{Discussion} \label{sec:6}

In this paper, we propose ReHarness for agentic continual learning beyond forward-only harness evolution.
ReHarness complements forward read and write with backward revision, using downstream feedback to revise past reflections while grounding new content in their source trajectories.
Across three continual learning benchmarks and three LLM backbones, ReHarness consistently outperforms strong baselines, with statistically significant gains across challenging tasks.
Further analyses show that ReHarness produces source-grounded and meaningful revisions, corrects overgeneralized or outdated reflections under distribution shifts, and improves stability and plasticity.
Together, our results highlight recurrent harness adaptation as a promising direction for continually refining accumulated experience under a strictly streaming setting.
Despite these gains, ReHarness underperforms on exploitable poker in CL-Bench, where downstream rewards provide noisy signals for backward revision.
Moreover, while ReHarness is more token-efficient than the strongest full-history raw-context baseline, it still stores raw trajectories for future revision, incurring additional storage overhead.
We aim to address these limitations in future work.

\clearpage

\bibliographystyle{plainnat}
\bibliography{main}

\clearpage
\beginappendix
\section{Appendix} \label{app:a}
\begin{figure}[htbp]
    \centering
    \includegraphics[width=0.9\textwidth]{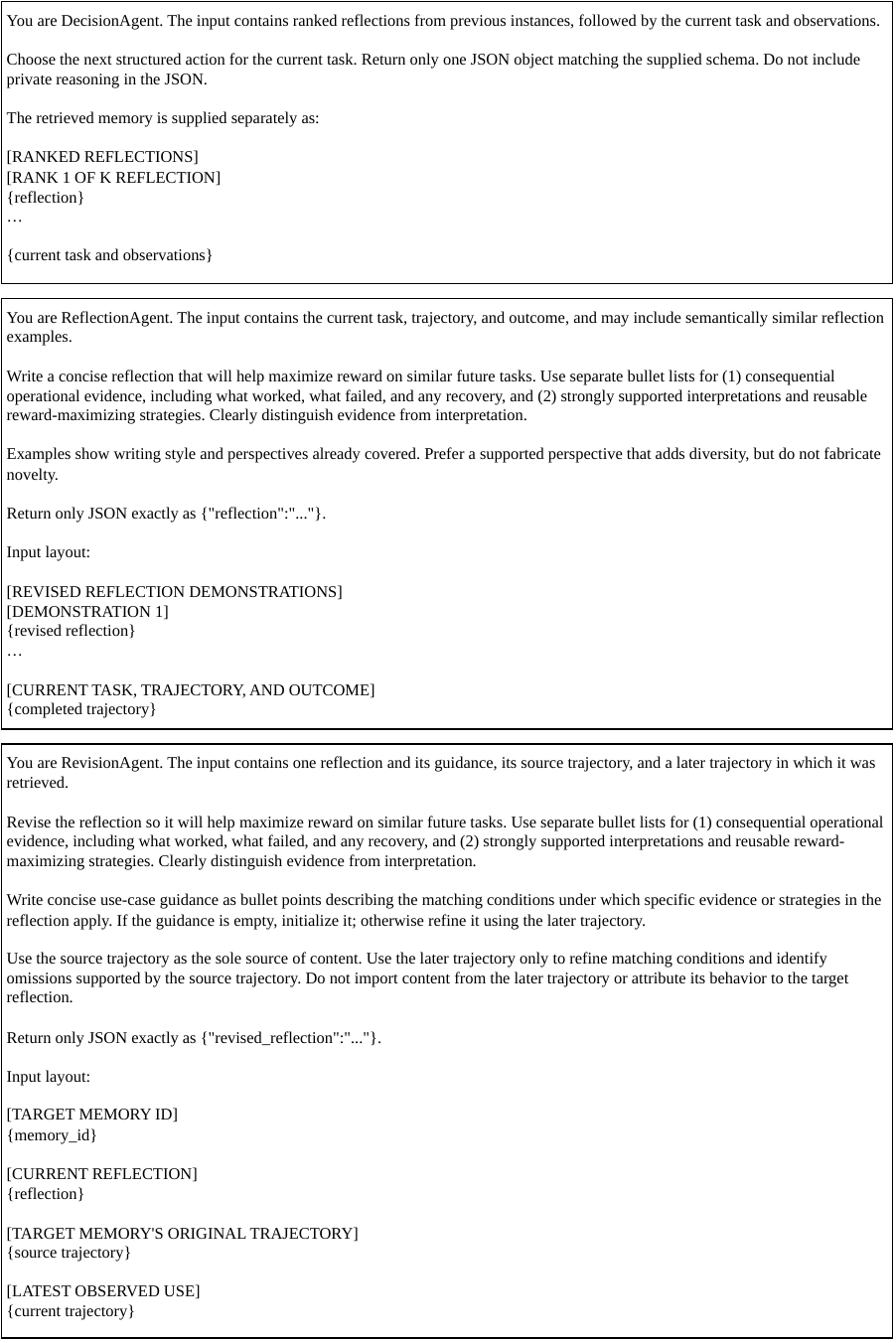}
    \caption{Prompts used by the decision (top), memory (middle), and revision (bottom) agents in ReHarness.}
    \label{fig:prompt}
\end{figure}

\begin{figure}[htbp]
    \centering
    \includegraphics[width=0.9\textwidth]{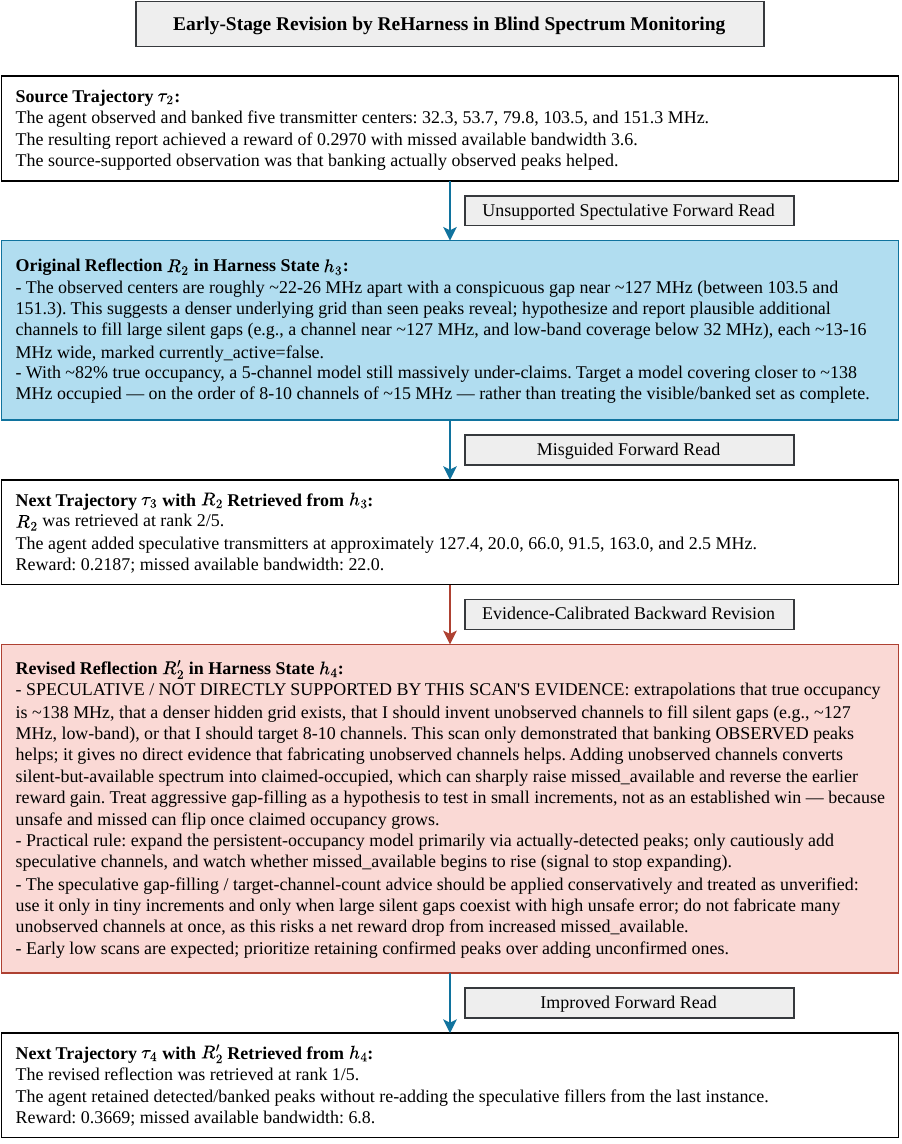}
    \caption{Full qualitative revision example for blind spectrum monitoring from Figure~\ref{fig:example} (a).}
    \label{fig:spectrum}
\end{figure}

\begin{figure}[htbp]
    \centering
    \includegraphics[width=0.9\textwidth]{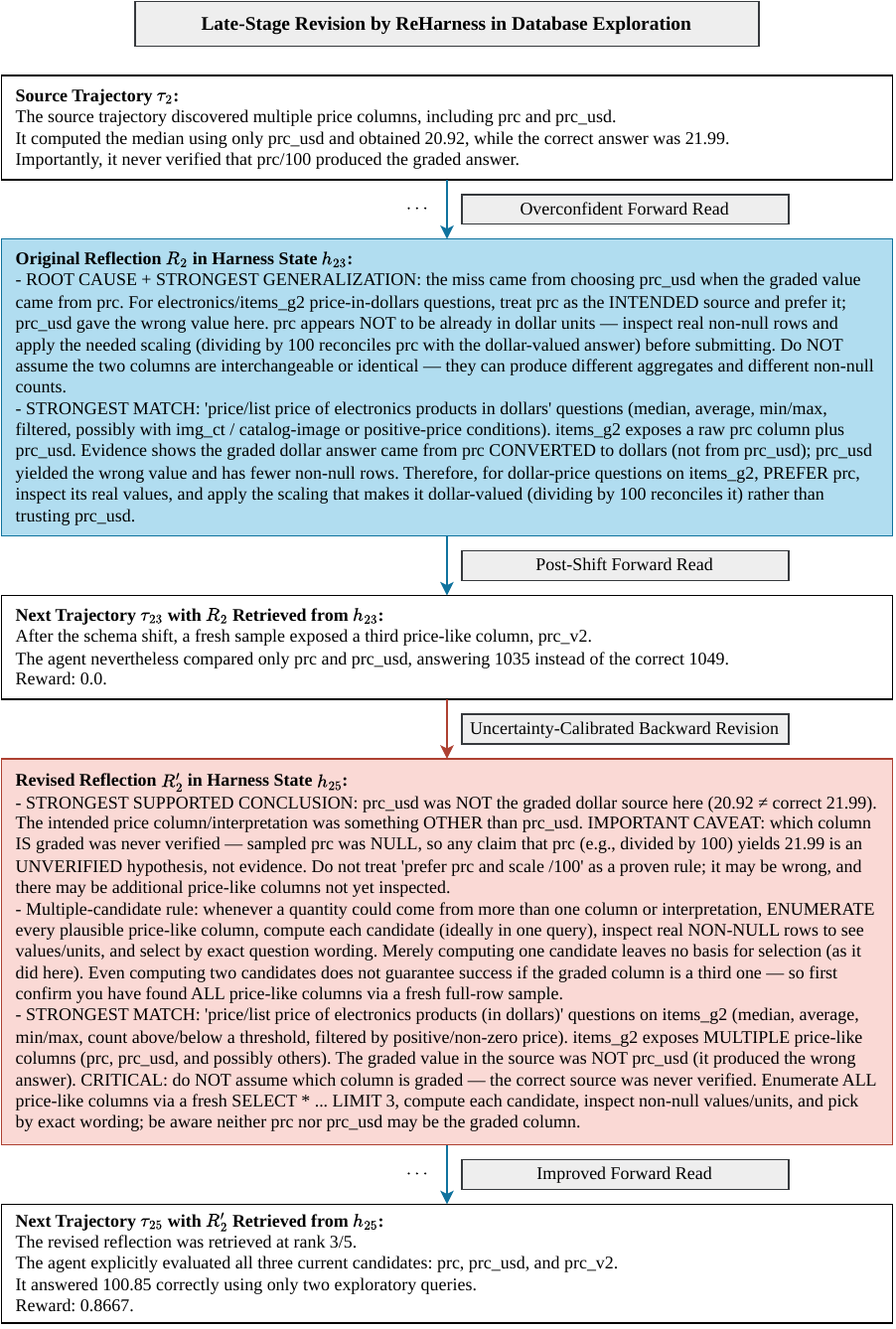}
    \caption{Full qualitative revision example for database exploration from Figure~\ref{fig:example} (b).}
    \label{fig:database}
\end{figure}

We report task-level normalized gains for each LLM on CL-Bench, MemoryBench, and MemoryArena in Tables~\ref{tab:clbench},~\ref{tab:memorybench}, and~\ref{tab:memoryarena}, respectively.
We provide the prompts used by the decision agent $\pi_D$, memory agent $\pi_M$, and revision agent $\pi_R$ in ReHarness in Figure~\ref{fig:prompt}.
We also provide the full qualitative revision examples discussed in \S\ref{sec:5.3} for blind spectrum monitoring and database exploration in Figures~\ref{fig:spectrum} and~\ref{fig:database}, respectively.

\begin{table}[htbp]
    \caption{Normalized gain for each LLM on each CL-Bench task, with the highest value in bold.}
    \centering
\resizebox{\textwidth}{!}{
    \begin{tabular}{c|cccccc|c}
        \toprule
            \textbf{Method} & \textbf{Cohort} & \textbf{Codebase} & \textbf{Database} & \textbf{Sales} & \textbf{Spectrum} & \textbf{Poker} & \textbf{Average} \\
        \midrule
        \midrule
        	\multicolumn{8}{c}{\textbf{LLM:} Claude Opus 4.8} \\
        \midrule
        	\textbf{Mem0} & $-0.045 \pm 0.014$ & $0.408 \pm 0.022$ & $-0.016 \pm 0.014$ & $0.284 \pm 0.011$ & $0.000 \pm 0.018$ & $-0.013 \pm 0.002$ & $0.103 \pm 0.006$ \\
        	\textbf{RBank} & $-0.065 \pm 0.002$ & $0.238 \pm 0.007$ & $-0.119 \pm 0.006$ & $0.376 \pm 0.011$ & $0.000 \pm 0.001$ & $0.005 \pm 0.003$ & $0.073 \pm 0.002$ \\
        	\textbf{DoT} & $-0.074 \pm 0.002$ & $\mathbf{0.499 \pm 0.009}$ & $0.056 \pm 0.006$ & $0.252 \pm 0.026$ & $0.033 \pm 0.004$ & $-0.017 \pm 0.001$ & $0.125 \pm 0.007$ \\
        	\textbf{ACE} & $\mathbf{-0.015 \pm 0.004}$ & $0.231 \pm 0.004$ & $-0.005 \pm 0.004$ & $0.051 \pm 0.006$ & $0.000 \pm 0.001$ & $-0.010 \pm 0.003$ & $0.042 \pm 0.001$ \\
        	\textbf{GEPA-Online} & $-0.028 \pm 0.006$ & $0.238 \pm 0.023$ & $0.040 \pm 0.014$ & $0.171 \pm 0.004$ & $-0.023 \pm 0.004$ & $\mathbf{0.056 \pm 0.002}$ & $0.076 \pm 0.008$ \\
        	\textbf{Top-K Raw} & $-0.119 \pm 0.010$ & $0.179 \pm 0.006$ & $0.371 \pm 0.005$ & $0.267 \pm 0.020$ & $\mathbf{0.275 \pm 0.003}$ & $0.014 \pm 0.005$ & $0.165 \pm 0.003$ \\
        	\textbf{Full Raw} & $-0.047 \pm 0.002$ & $0.364 \pm 0.010$ & $\mathbf{0.406 \pm 0.004}$ & $0.229 \pm 0.013$ & $0.216 \pm 0.007$ & $-0.020 \pm 0.012$ & $0.191 \pm 0.004$ \\
        	\textbf{ReHarness} & $-0.019 \pm 0.012$ & $0.378 \pm 0.011$ & $0.361 \pm 0.004$ & $\mathbf{0.408 \pm 0.018}$ & $0.162 \pm 0.014$ & $0.009 \pm 0.002$ & $\mathbf{0.217 \pm 0.006}$ \\
        \midrule
        	\multicolumn{8}{c}{\textbf{LLM:} GPT-5.5} \\
        \midrule
        	\textbf{Mem0} & $-0.047 \pm 0.006$ & $0.402 \pm 0.019$ & $0.349 \pm 0.012$ & $0.365 \pm 0.033$ & $0.233 \pm 0.014$ & $0.020 \pm 0.002$ & $0.220 \pm 0.006$ \\
        	\textbf{RBank} & $-0.029 \pm 0.003$ & $0.319 \pm 0.009$ & $-0.053 \pm 0.005$ & $0.068 \pm 0.005$ & $0.000 \pm 0.001$ & $0.087 \pm 0.004$ & $0.065 \pm 0.002$ \\
        	\textbf{DoT} & $-0.105 \pm 0.002$ & $0.228 \pm 0.010$ & $0.226 \pm 0.005$ & $0.078 \pm 0.016$ & $0.138 \pm 0.002$ & $-0.080 \pm 0.005$ & $0.081 \pm 0.005$ \\
        	\textbf{ACE} & $-0.109 \pm 0.003$ & $0.152 \pm 0.001$ & $0.118 \pm 0.006$ & $0.322 \pm 0.012$ & $0.000 \pm 0.001$ & $0.044 \pm 0.004$ & $0.088 \pm 0.003$ \\
        	\textbf{GEPA-Online} & $0.121 \pm 0.006$ & $\mathbf{0.694 \pm 0.014}$ & $0.307 \pm 0.011$ & $0.110 \pm 0.005$ & $0.124 \pm 0.009$ & $0.076 \pm 0.003$ & $0.239 \pm 0.008$ \\
        	\textbf{Top-K Raw} & $0.011 \pm 0.012$ & $0.636 \pm 0.020$ & $0.465 \pm 0.008$ & $-0.002 \pm 0.030$ & $0.302 \pm 0.005$ & $0.196 \pm 0.005$ & $0.268 \pm 0.007$ \\
        	\textbf{Full Raw} & $-0.068 \pm 0.002$ & $0.638 \pm 0.008$ & $0.482 \pm 0.001$ & $-0.022 \pm 0.013$ & $\mathbf{0.344 \pm 0.004}$ & $\mathbf{0.421 \pm 0.017}$ & $0.299 \pm 0.003$ \\
        	\textbf{ReHarness} & $\mathbf{0.234 \pm 0.014}$ & $0.500 \pm 0.009$ & $\mathbf{0.515 \pm 0.009}$ & $\mathbf{0.437 \pm 0.011}$ & $0.198 \pm 0.017$ & $0.016 \pm 0.001$ & $\mathbf{0.317 \pm 0.004}$ \\
        \midrule
        	\multicolumn{8}{c}{\textbf{LLM:} Gemini 3.7 Flash} \\
        \midrule
        	\textbf{Mem0} & $0.190 \pm 0.007$ & $0.066 \pm 0.010$ & $0.271 \pm 0.010$ & $-0.337 \pm 0.014$ & $0.357 \pm 0.007$ & $0.003 \pm 0.002$ & $0.092 \pm 0.005$ \\
        	\textbf{RBank} & $-0.014 \pm 0.003$ & $0.058 \pm 0.010$ & $0.017 \pm 0.007$ & $0.182 \pm 0.005$ & $0.000 \pm 0.001$ & $-0.003 \pm 0.005$ & $0.040 \pm 0.002$ \\
        	\textbf{DoT} & $-0.124 \pm 0.002$ & $0.247 \pm 0.013$ & $0.077 \pm 0.005$ & $-0.249 \pm 0.011$ & $0.014 \pm 0.006$ & $-0.011 \pm 0.003$ & $-0.008 \pm 0.004$ \\
        	\textbf{ACE} & $-0.047 \pm 0.003$ & $0.129 \pm 0.003$ & $0.089 \pm 0.004$ & $0.085 \pm 0.014$ & $0.000 \pm 0.001$ & $-0.038 \pm 0.004$ & $0.036 \pm 0.003$ \\
        	\textbf{GEPA-Online} & $0.039 \pm 0.004$ & $0.286 \pm 0.005$ & $\mathbf{0.459 \pm 0.009}$ & $0.041 \pm 0.004$ & $-0.043 \pm 0.005$ & $0.012 \pm 0.004$ & $0.132 \pm 0.007$ \\
        	\textbf{Top-K Raw} & $\mathbf{0.209 \pm 0.018}$ & $\mathbf{0.701 \pm 0.012}$ & $0.365 \pm 0.007$ & $-0.356 \pm 0.029$ & $0.222 \pm 0.003$ & $-0.004 \pm 0.009$ & $0.190 \pm 0.006$ \\
        	\textbf{Full Raw} & $-0.036 \pm 0.002$ & $0.440 \pm 0.009$ & $0.426 \pm 0.003$ & $\mathbf{0.343 \pm 0.002}$ & $0.160 \pm 0.007$ & $\mathbf{0.070 \pm 0.011}$ & $\mathbf{0.234 \pm 0.003}$ \\
        	\textbf{ReHarness} & $0.044 \pm 0.010$ & $0.610 \pm 0.004$ & $0.432 \pm 0.007$ & $-0.141 \pm 0.024$ & $\mathbf{0.458 \pm 0.009}$ & $-0.005 \pm 0.002$ & $0.233 \pm 0.003$ \\
        \bottomrule
    \end{tabular}
}
    \label{tab:clbench}
\end{table}

\begin{table}[htbp]
    \caption{Normalized gain for each LLM on each MemoryBench task, with the highest value in bold.}
    \centering
\resizebox{\textwidth}{!}{
    \begin{tabular}{c|ccccc|c}
        \toprule
            \textbf{Method} & \textbf{HB-Academic} & \textbf{HB-Creative} & \textbf{WB-Academic} & \textbf{WB-Creative} & \textbf{WB-Politics} & \textbf{Average} \\
        \midrule
        \midrule
        	\multicolumn{7}{c}{\textbf{LLM:} Claude Opus 4.8} \\
        \midrule
        	\textbf{Mem0} & $0.073 \pm 0.004$ & $\mathbf{0.182 \pm 0.009}$ & $0.000 \pm 0.004$ & $0.053 \pm 0.003$ & $-0.065 \pm 0.009$ & $0.049 \pm 0.002$ \\
        	\textbf{RBank} & $-0.055 \pm 0.004$ & $0.111 \pm 0.004$ & $0.026 \pm 0.002$ & $0.033 \pm 0.003$ & $0.000 \pm 0.003$ & $0.023 \pm 0.002$ \\
        	\textbf{DoT} & $0.077 \pm 0.002$ & $0.179 \pm 0.004$ & $-0.013 \pm 0.002$ & $0.047 \pm 0.003$ & $-0.026 \pm 0.003$ & $0.053 \pm 0.003$ \\
        	\textbf{ACE} & $0.115 \pm 0.004$ & $0.134 \pm 0.001$ & $0.064 \pm 0.007$ & $0.113 \pm 0.005$ & $0.000 \pm 0.004$ & $0.085 \pm 0.004$ \\
        	\textbf{GEPA-Online} & $0.166 \pm 0.005$ & $0.132 \pm 0.002$ & $0.090 \pm 0.029$ & $\mathbf{0.173 \pm 0.003}$ & $\mathbf{0.130 \pm 0.006}$ & $0.138 \pm 0.013$ \\
        	\textbf{Top-K Raw} & $0.118 \pm 0.003$ & $-0.008 \pm 0.004$ & $0.026 \pm 0.002$ & $0.040 \pm 0.002$ & $0.000 \pm 0.002$ & $0.035 \pm 0.001$ \\
        	\textbf{Full Raw} & $0.122 \pm 0.003$ & $0.167 \pm 0.003$ & $0.000 \pm 0.003$ & $0.080 \pm 0.005$ & $-0.052 \pm 0.002$ & $0.063 \pm 0.002$ \\
        	\textbf{ReHarness} & $\mathbf{0.199 \pm 0.009}$ & $0.158 \pm 0.002$ & $\mathbf{0.141 \pm 0.006}$ & $\mathbf{0.173 \pm 0.006}$ & $\mathbf{0.130 \pm 0.005}$ & $\mathbf{0.160 \pm 0.003}$ \\
        \midrule
        	\multicolumn{7}{c}{\textbf{LLM:} GPT-5.5} \\
        \midrule
        	\textbf{Mem0} & $0.094 \pm 0.003$ & $0.070 \pm 0.008$ & $-0.016 \pm 0.005$ & $0.000 \pm 0.003$ & $-0.066 \pm 0.005$ & $0.016 \pm 0.003$ \\
        	\textbf{RBank} & $-0.009 \pm 0.005$ & $0.101 \pm 0.006$ & $-0.032 \pm 0.002$ & $0.009 \pm 0.003$ & $-0.049 \pm 0.003$ & $0.004 \pm 0.003$ \\
        	\textbf{DoT} & $0.100 \pm 0.003$ & $0.092 \pm 0.002$ & $-0.016 \pm 0.004$ & $0.009 \pm 0.004$ & $-0.049 \pm 0.003$ & $0.027 \pm 0.002$ \\
        	\textbf{ACE} & $-0.008 \pm 0.002$ & $0.071 \pm 0.004$ & $-0.063 \pm 0.005$ & $0.009 \pm 0.003$ & $-0.098 \pm 0.004$ & $-0.018 \pm 0.004$ \\
        	\textbf{GEPA-Online} & $0.163 \pm 0.007$ & $0.139 \pm 0.002$ & $0.032 \pm 0.007$ & $0.000 \pm 0.009$ & $-0.016 \pm 0.006$ & $0.064 \pm 0.006$ \\
        	\textbf{Top-K Raw} & $0.169 \pm 0.001$ & $0.051 \pm 0.007$ & $-0.032 \pm 0.003$ & $0.035 \pm 0.004$ & $-0.016 \pm 0.002$ & $0.041 \pm 0.001$ \\
        	\textbf{Full Raw} & $0.157 \pm 0.007$ & $0.102 \pm 0.003$ & $-0.016 \pm 0.003$ & $-0.009 \pm 0.006$ & $-0.115 \pm 0.004$ & $0.024 \pm 0.003$ \\
        	\textbf{ReHarness} & $\mathbf{0.215 \pm 0.010}$ & $\mathbf{0.172 \pm 0.002}$ & $\mathbf{0.048 \pm 0.005}$ & $\mathbf{0.044 \pm 0.010}$ & $\mathbf{0.000 \pm 0.007}$ & $\mathbf{0.096 \pm 0.006}$ \\
        \midrule
        	\multicolumn{7}{c}{\textbf{LLM:} Gemini 3.7 Flash} \\
        \midrule
        	\textbf{Mem0} & $\mathbf{0.135 \pm 0.002}$ & $0.026 \pm 0.010$ & $\mathbf{0.076 \pm 0.004}$ & $-0.045 \pm 0.004$ & $\mathbf{0.071 \pm 0.001}$ & $0.053 \pm 0.002$ \\
        	\textbf{RBank} & $0.031 \pm 0.003$ & $\mathbf{0.193 \pm 0.004}$ & $0.000 \pm 0.002$ & $-0.025 \pm 0.004$ & $0.024 \pm 0.004$ & $0.045 \pm 0.003$ \\
        	\textbf{DoT} & $0.045 \pm 0.002$ & $0.094 \pm 0.006$ & $-0.076 \pm 0.001$ & $-0.025 \pm 0.003$ & $0.012 \pm 0.002$ & $0.010 \pm 0.002$ \\
        	\textbf{ACE} & $0.098 \pm 0.004$ & $0.069 \pm 0.002$ & $-0.013 \pm 0.003$ & $0.000 \pm 0.004$ & $-0.012 \pm 0.004$ & $0.028 \pm 0.005$ \\
        	\textbf{GEPA-Online} & $-0.062 \pm 0.010$ & $0.119 \pm 0.002$ & $-0.772 \pm 0.010$ & $\mathbf{0.032 \pm 0.005}$ & $0.012 \pm 0.007$ & $-0.134 \pm 0.010$ \\
        	\textbf{Top-K Raw} & $0.113 \pm 0.002$ & $0.131 \pm 0.006$ & $-0.013 \pm 0.002$ & $-0.032 \pm 0.003$ & $-0.012 \pm 0.002$ & $0.037 \pm 0.001$ \\
        	\textbf{Full Raw} & $-0.032 \pm 0.006$ & $0.153 \pm 0.003$ & $-0.063 \pm 0.002$ & $-0.064 \pm 0.005$ & $-0.060 \pm 0.004$ & $-0.013 \pm 0.003$ \\
        	\textbf{ReHarness} & $0.040 \pm 0.009$ & $0.173 \pm 0.002$ & $0.013 \pm 0.004$ & $\mathbf{0.032 \pm 0.002}$ & $0.060 \pm 0.004$ & $\mathbf{0.064 \pm 0.003}$ \\
        \bottomrule
    \end{tabular}
}
    \label{tab:memorybench}
\end{table}

\begin{table}[htbp]
    \caption{Normalized gain for each LLM on each MemoryArena task, with the highest value in bold.}
    \centering
\resizebox{\textwidth}{!}{
    \begin{tabular}{c|ccccc|c}
        \toprule
            \textbf{Method} & \textbf{Travel} & \textbf{Shopping} & \textbf{Search} & \textbf{Math} & \textbf{Physics} & \textbf{Average} \\
        \midrule
        \midrule
        	\multicolumn{7}{c}{\textbf{LLM:} Claude Opus 4.8} \\
        \midrule
        	\textbf{Mem0} & $\mathbf{0.451 \pm 0.022}$ & $0.029 \pm 0.028$ & $-0.057 \pm 0.003$ & $0.016 \pm 0.002$ & $0.024 \pm 0.014$ & $0.093 \pm 0.005$ \\
        	\textbf{RBank} & $0.037 \pm 0.003$ & $0.043 \pm 0.002$ & $0.019 \pm 0.002$ & $-0.020 \pm 0.003$ & $-0.010 \pm 0.004$ & $0.014 \pm 0.001$ \\
        	\textbf{DoT} & $0.085 \pm 0.004$ & $0.121 \pm 0.010$ & $0.124 \pm 0.002$ & $0.013 \pm 0.004$ & $0.300 \pm 0.008$ & $0.129 \pm 0.008$ \\
        	\textbf{ACE} & $-0.002 \pm 0.001$ & $0.079 \pm 0.001$ & $0.048 \pm 0.004$ & $0.012 \pm 0.002$ & $0.089 \pm 0.007$ & $0.045 \pm 0.003$ \\
        	\textbf{GEPA-Online} & $0.009 \pm 0.004$ & $0.219 \pm 0.014$ & $0.105 \pm 0.007$ & $0.077 \pm 0.002$ & $0.218 \pm 0.007$ & $0.126 \pm 0.005$ \\
        	\textbf{Top-K Raw} & $0.292 \pm 0.022$ & $0.138 \pm 0.007$ & $-0.514 \pm 0.011$ & $0.148 \pm 0.001$ & $\mathbf{0.663 \pm 0.017}$ & $0.145 \pm 0.007$ \\
        	\textbf{Full Raw} & $0.274 \pm 0.008$ & $0.179 \pm 0.012$ & $-0.819 \pm 0.027$ & $\mathbf{0.192 \pm 0.002}$ & $0.582 \pm 0.005$ & $0.082 \pm 0.006$ \\
        	\textbf{ReHarness} & $0.443 \pm 0.022$ & $\mathbf{0.221 \pm 0.008}$ & $\mathbf{0.210 \pm 0.007}$ & $0.138 \pm 0.003$ & $0.511 \pm 0.007$ & $\mathbf{0.305 \pm 0.003}$ \\
        \midrule
        	\multicolumn{7}{c}{\textbf{LLM:} GPT-5.5} \\
        \midrule
        	\textbf{Mem0} & $0.771 \pm 0.013$ & $0.466 \pm 0.024$ & $0.000 \pm 0.003$ & $0.103 \pm 0.002$ & $0.271 \pm 0.008$ & $0.322 \pm 0.010$ \\
        	\textbf{RBank} & $0.000 \pm 0.003$ & $0.011 \pm 0.003$ & $0.014 \pm 0.002$ & $0.010 \pm 0.003$ & $0.048 \pm 0.003$ & $0.017 \pm 0.001$ \\
        	\textbf{DoT} & $0.176 \pm 0.007$ & $0.383 \pm 0.006$ & $0.060 \pm 0.003$ & $0.119 \pm 0.006$ & $0.155 \pm 0.009$ & $0.179 \pm 0.005$ \\
        	\textbf{ACE} & $-0.012 \pm 0.001$ & $0.059 \pm 0.002$ & $0.009 \pm 0.004$ & $0.015 \pm 0.002$ & $-0.051 \pm 0.007$ & $0.004 \pm 0.003$ \\
        	\textbf{GEPA-Online} & $-0.149 \pm 0.005$ & $0.327 \pm 0.008$ & $\mathbf{0.321 \pm 0.003}$ & $-0.011 \pm 0.006$ & $0.155 \pm 0.007$ & $0.129 \pm 0.002$ \\
        	\textbf{Top-K Raw} & $0.781 \pm 0.002$ & $0.425 \pm 0.013$ & $0.023 \pm 0.020$ & $0.150 \pm 0.001$ & $0.319 \pm 0.015$ & $0.340 \pm 0.009$ \\
        	\textbf{Full Raw} & $0.787 \pm 0.023$ & $\mathbf{0.471 \pm 0.010}$ & $0.014 \pm 0.022$ & $\mathbf{0.168 \pm 0.003}$ & $0.350 \pm 0.011$ & $\mathbf{0.358 \pm 0.009}$ \\
        	\textbf{ReHarness} & $\mathbf{0.816 \pm 0.024}$ & $0.407 \pm 0.008$ & $0.101 \pm 0.007$ & $0.115 \pm 0.004$ & $\mathbf{0.351 \pm 0.010}$ & $\mathbf{0.358 \pm 0.006}$ \\
        \midrule
        	\multicolumn{7}{c}{\textbf{LLM:} Gemini 3.7 Flash} \\
        \midrule
        	\textbf{Mem0} & $0.904 \pm 0.010$ & $0.005 \pm 0.031$ & $-0.030 \pm 0.003$ & $0.089 \pm 0.003$ & $0.209 \pm 0.002$ & $0.235 \pm 0.008$ \\
        	\textbf{RBank} & $-0.005 \pm 0.002$ & $0.037 \pm 0.002$ & $0.045 \pm 0.001$ & $-0.031 \pm 0.003$ & $0.069 \pm 0.004$ & $0.023 \pm 0.001$ \\
        	\textbf{DoT} & $0.009 \pm 0.004$ & $0.072 \pm 0.017$ & $0.082 \pm 0.002$ & $0.013 \pm 0.005$ & $0.035 \pm 0.007$ & $0.042 \pm 0.004$ \\
        	\textbf{ACE} & $-0.009 \pm 0.001$ & $0.058 \pm 0.001$ & $0.090 \pm 0.002$ & $0.048 \pm 0.002$ & $0.180 \pm 0.008$ & $0.073 \pm 0.004$ \\
        	\textbf{GEPA-Online} & $0.000 \pm 0.007$ & $0.070 \pm 0.012$ & $0.201 \pm 0.004$ & $-0.008 \pm 0.003$ & $0.028 \pm 0.008$ & $0.058 \pm 0.004$ \\
        	\textbf{Top-K Raw} & $0.940 \pm 0.014$ & $0.221 \pm 0.013$ & $-0.269 \pm 0.014$ & $\mathbf{0.156 \pm 0.001}$ & $0.407 \pm 0.013$ & $0.291 \pm 0.007$ \\
        	\textbf{Full Raw} & $\mathbf{0.945 \pm 0.017}$ & $0.182 \pm 0.008$ & $-0.537 \pm 0.016$ & $0.142 \pm 0.002$ & $0.452 \pm 0.010$ & $0.237 \pm 0.010$ \\
        	\textbf{ReHarness} & $0.930 \pm 0.021$ & $\mathbf{0.226 \pm 0.009}$ & $\mathbf{0.291 \pm 0.005}$ & $0.086 \pm 0.002$ & $\mathbf{0.508 \pm 0.004}$ & $\mathbf{0.408 \pm 0.003}$ \\
        \bottomrule
    \end{tabular}
}
    \label{tab:memoryarena}
\end{table}

\end{document}